\documentclass[conference]{IEEEtran}
\ifCLASSINFOpdf
   \usepackage[pdftex]{graphicx}
   \DeclareGraphicsExtensions{.pdf,.jpeg,.png}
\else
   \usepackage[dvips]{graphicx}
   \DeclareGraphicsExtensions{.eps}
\fi
\usepackage{amsmath}

\usepackage{wasysym}

\def\BibTeX{{\rm B\kern-.05em{\sc i\kern-.025em b}\kern-.08em
    T\kern-.1667em\lower.7ex\hbox{E}\kern-.125emX}}

\usepackage{xcolor}

\usepackage[flushleft]{threeparttable}

\usepackage{tikz}
\newcommand\copyrighttext{%
  \footnotesize \textsuperscript{\textcopyright} 2021 IEEE. Personal use of this material is permitted.  Permission from IEEE must be obtained for all other uses, in any current or future media, including reprinting/republishing this material for advertising or promotional purposes, creating new collective works, for resale or redistribution to servers or lists, or reuse of any copyrighted component of this work in other works.}
\newcommand\copyrightnotice{%
\begin{tikzpicture}[remember picture,overlay]
\node[anchor=south,yshift=0pt] at (current page.south) {\fbox{\parbox{\dimexpr\textwidth-\fboxsep-\fboxrule\relax}{\copyrighttext}}};
\end{tikzpicture}%
}

\begin{document}
\bstctlcite{BibControl}
%
\title{Design of a user-friendly control system for planetary rovers with CPS feature}

\makeatletter
\newcommand{\linebreakand}{%
  \end{@IEEEauthorhalign}
  \hfill\mbox{}\par
  \mbox{}\hfill\begin{@IEEEauthorhalign}
}
\makeatother

\author{

\IEEEauthorblockN{Sebastiano Chiodini}
\IEEEauthorblockA{\textit{CISAS "Giuseppe Colombo"} \\
\textit{University of Padova}\\
Padova, Italy\\
sebastiano.chiodini@unipd.it}
\and

\IEEEauthorblockN{Riccardo Giubilato}
\IEEEauthorblockA{\textit{Institute of Robotics and Mechatronics} \\
\textit{DLR}\\
Wessling, Germany\\
riccardo.giubilato@dlr.de
}
\and

\IEEEauthorblockN{Marco Pertile}
\IEEEauthorblockA{\textit{Dept. of Industrial Engineering} \\
\textit{University of Padova}\\
Padova, Italy\\
marco.pertile@unipd.it}
\linebreakand 

\IEEEauthorblockN{Annarita Tedesco}
\IEEEauthorblockA{\textit{IMS  Laboratory} \\
\textit{University of Bordeaux}\\
Bordeaux, France\\
annarita.tedesco@ims-bordeaux.fr}
\and

\IEEEauthorblockN{Domenico Accardo}
\IEEEauthorblockA{\textit{Dept. of Industrial Engineering} \\
\textit{University of Napoli Federico II}\\
Napoli, Italy\\
domenico.accardo@unina.it}
\and

\IEEEauthorblockN{Stefano Debei}
\IEEEauthorblockA{\textit{Dept. of Industrial Engineering} \\
\textit{University of Padova}\\
Padova, Italy\\
stefano.debei@unipd.it}

}

\maketitle

\copyrightnotice{}

\begin{abstract}
In this paper, we present a user-friendly planetary rover’s control system for low latency surface telerobotic. Thanks to the proposed system, an operator can comfortably give commands through the control base station to a rover using commercially available off-the-shelf (COTS) joysticks or by command sequencing with interactive monitoring on the sensed map of the environment. During operations, high situational awareness is made possible thanks to 3D map visualization. The map of the environment is built on the on-board computer by processing the rover's camera images with a visual Simultaneous Localization and Mapping (SLAM) algorithm. It is transmitted via Wi-Fi and displayed on the control base station screen in near real-time. The navigation stack takes as input the visual SLAM data to build a cost map to find the minimum cost path. By interacting with the virtual map, the rover exhibits properties of a Cyber Physical System (CPS) for its self-awareness capabilities. The software architecture is based on the Robot Operative System (ROS) middleware. The system design and the preliminary field test results are shown in the paper.
\end{abstract}


%
\IEEEpeerreviewmaketitle

\section{Introduction}
Planetary rovers are measuring system machines, their cameras, as well as being indispensable for rover's operations planning, provide on-site measurements of the geological context. Up to now, due to the latency in uplinking commands and downlinking telemetry, operations teams planify rover navigation targets on daily bases and rely on rover's fault protection, autonomous navigation, and visual odometry software to keep the rover safe during drives \cite{rankin2021}.

The possibility to control a rover by astronauts located in a control station on the planet's surface or orbiting the planet paves the way to new explorations capabilities. Indeed, thanks to low latency surface telerobotics, astronauts can be telepresent on the planetary surface in a highly productive manner. As an example, in future NASA and ESA human exploration missions, low latency rover control is foreseen from the crewed module of the Lunar Orbital Platform - Gateway (LOP-G) to carry out astronaut assisted sample return and deployment/construction of a low-frequency radio telescope array \cite{burns2019}. \cite{bualat2013} reports the campaign results of how astronauts in the International Space Station (ISS) have remotely operated a planetary rover for telescope deployment on Earth. The test has been performed by the astronauts using a Space Station Computer (Lenovo Thinkpad laptop), supervisory control (command sequencing with interactive monitoring), teleoperation (discrete commanding), and Ku-band satellite communications to remotely operate a rover. In \cite{sonsalla2017field} the authors present the remote operation from a control station located in Bremen, Germany, of a rover located in Mars analogue terrain in the desert of Utah, USA. In \cite{Schuster2020arches} is described the ARCHES demonstration missions, where an heterogeneous robotic team will be deployed in the summer of 2021 at a Moon-analogue site on the volcano Mt. Etna on Sicily, Italy.

We can consider the latency of a signal to control a teleoperated robot to be low, and thus speak about low-latency telerobotics, if the delay of the signal is small enough not to make teleoperation of a robot unpleasant or impossible \cite{lester2011human}.  Whereas trained surgeons can even do precision surgical tasks with 500 ms of latency, the latency of a signal between the surface of the Moon and Earth-Moon L1 or L2 is 410 ms, thus low latency teleoperation of rovers on the Moon surface seems to be feasible.  \cite{mellinkoff2018quantifying} quantifies the operational
video conditions required for effective exploration for planetary surface operations. 

Low latency  telerobotics  have been carried out also for Unmanned Ground Vehicle (UGV) control over 4G network, in \cite{manzi2016} near real-time robot control with video stream feedback over a 4G network has been performed. The development of 5G will surely open up new opportunities for even more immersive experiences for rover and UGV control \cite{civerchia2020remote}. In the near future, phased array technology could play a significant role in providing high data throughput links from the lunar surface to the lunar relay \cite{Nessel2020}.

Despite the ability to control these platforms in near real-time, autonomous navigation capabilities such as egomotion estimation and hazard avoidance are a fundamental asset. As communication delays up to a few seconds are still present and the system may be subject to communication interruptions. Simultaneous Localization and Mapping (SLAM) is a valuable approach for accurate robot self-localization and map building, an insight of the topic can be found in \cite{durrant-Whyte2006}. By interacting with a virtual map, the rover exhibits Cyber Physical System (CPS) feature, it can self-adjust its path based on the sensed 3D map of the environment \cite{lee2015}. Moreover, the 3D map of the environment can be used to increase the situational awareness of the operator, and for measurement and reconstruction of the context of operations.

On such bases we present a user-friendly planetary rover’s control system for low latency surface telerobotics. Compared to the solution presented by \cite{mellinkoff2018quantifying} where the control of the rover is based on joystick input and video feedback, the proposed system gives the possibility to set navigation goals that are reached autonomously and safely by the rover. The system is based on a decentralized architecture consisting of a control base station, an on-board computer (master) and a series of slave microcontrollers (Arduino), each running Robot Operating System (ROS) middleware \cite{quigley2009ros}. \cite{Snider2019} provides an interesting tutorial on how to set up a robot with ROS. This modular approach allows more flexibility in upgrading the rover hardware and its sensing elements. The goal was to develop a simple and intuitive control interface based on joystick commands for near-real time control, or by command sequencing with interactive monitoring on the surrounding environment map. Tests have been performed with MORPHEUS rover. MORPHEUS (Mars Operative Rover of Padova Engineering University Students) is a field robotic testbed under development at the University of Padova. It has been designed by a team of students as demonstrator for planetary robotic exploration technologies, like soil and rocks extraction and sampling, and autonomous navigation in unstructured environment. 

The work is divided as follows, in Section \ref{sec::design} we described the design and implementation of the platform, in Section \ref{sec::test} we report the preliminary test results and in Section \ref{sec::conclusiond} the conclusions of this work are reported.

\section{Design and Implementation}
\label{sec::design}

The employed platform is a six-wheel skid steering rover, which gives agility to the rover allowing from point turning to line driving with a reduced amount of actuators.  Each wheel has its own brushless motor and a transmission belt mechanism for the connection system. The configuration with three rockers, everyone equipped with two wheels, has been chosen to guarantee sufficient adherence to the soil. Rover dimensions are $820\times130\times1000$ mm, and the total mass is around 34 kg. Motor controllers are Maxon DEC Module 50/5 (Digital EC Controller), which is a 1-quadrant amplifier for controlling EC motors with Hall sensors with a maximum output of 250 watts. Motors are Maxon EC-max 30 $\diameter$ 30 mm, brushless, 60 Watt, with Hall sensors. The motors are coupled with a planetary gearhead with a 86 : 1 gear ratio.

\begin{figure}[!t]
\centering
\includegraphics[width=3.5in]{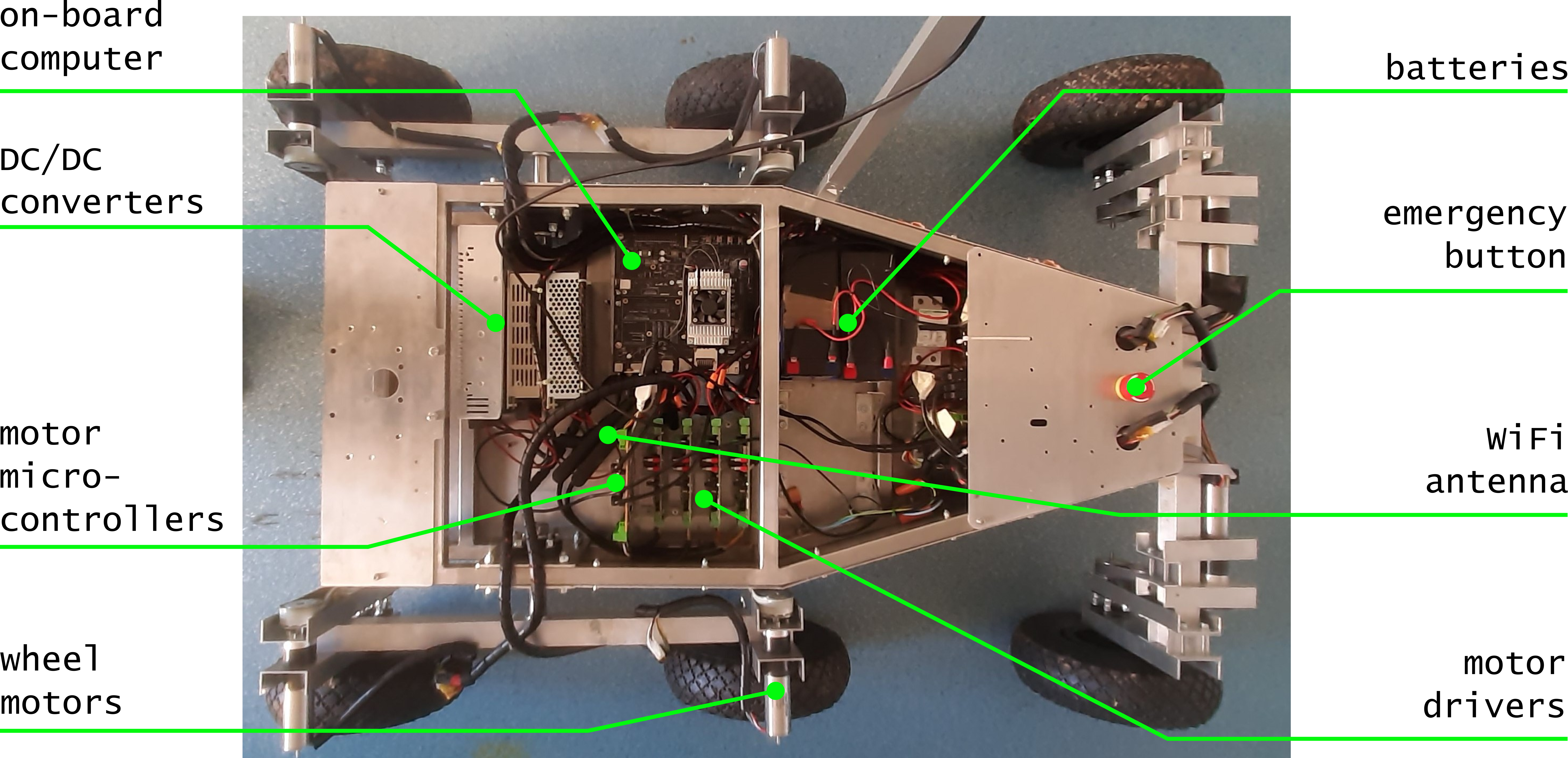}

\caption{Rover electronic control unit.}
\label{fig::rover_electronic}
\end{figure}

The rover on-board computer is an NVIDIA Jetson TX2 board with a hex-core ARMv8 CPU, and a 256-core Pascal GPU, and 8GB RAM. As an operating system (OS), we use a version of the 64-bit release of Ubuntu 16.04.6 LTS Xenial Xerus. The ground station consists of a separate laptop, also running a 16.04.6 LTS Xenial Xerus OS, and interacts with the rover’s Jetson TX2 boards through 5GHz 802.11ac Wi-Fi. A set of Arduino Nano, a small board based on the ATmega328 is used to convert the input from the on-board computer to the motor drivers. Each Arduino Nano microcontroller gives the inputs towards two motors, and is connected to the Jetson TX2 board by an USB cable. A picture of the platform and the main electronics components is shown in \figurename{\ref{fig::rover_electronic}}.

The rover is equipped with a ZED\footnote{https://www.stereolabs.com/} synchronized stereo camera equipped with wide angle lenses dedicated to the perception of the environment and operator visual feedback. Stereo images have been captured at 4 fps with a resolution of $1280\times720$ pixels.
The software is based on the ROS middleware, which is a flexible framework used for robotics research. The following paragraphs highlights how the proposed system is interconnected thanks to ROS, in \figurename{\ref{fig::system}} it is shown a sketch of the interconnected system.

\begin{figure}[!b]
\centering
\includegraphics[width=3in]{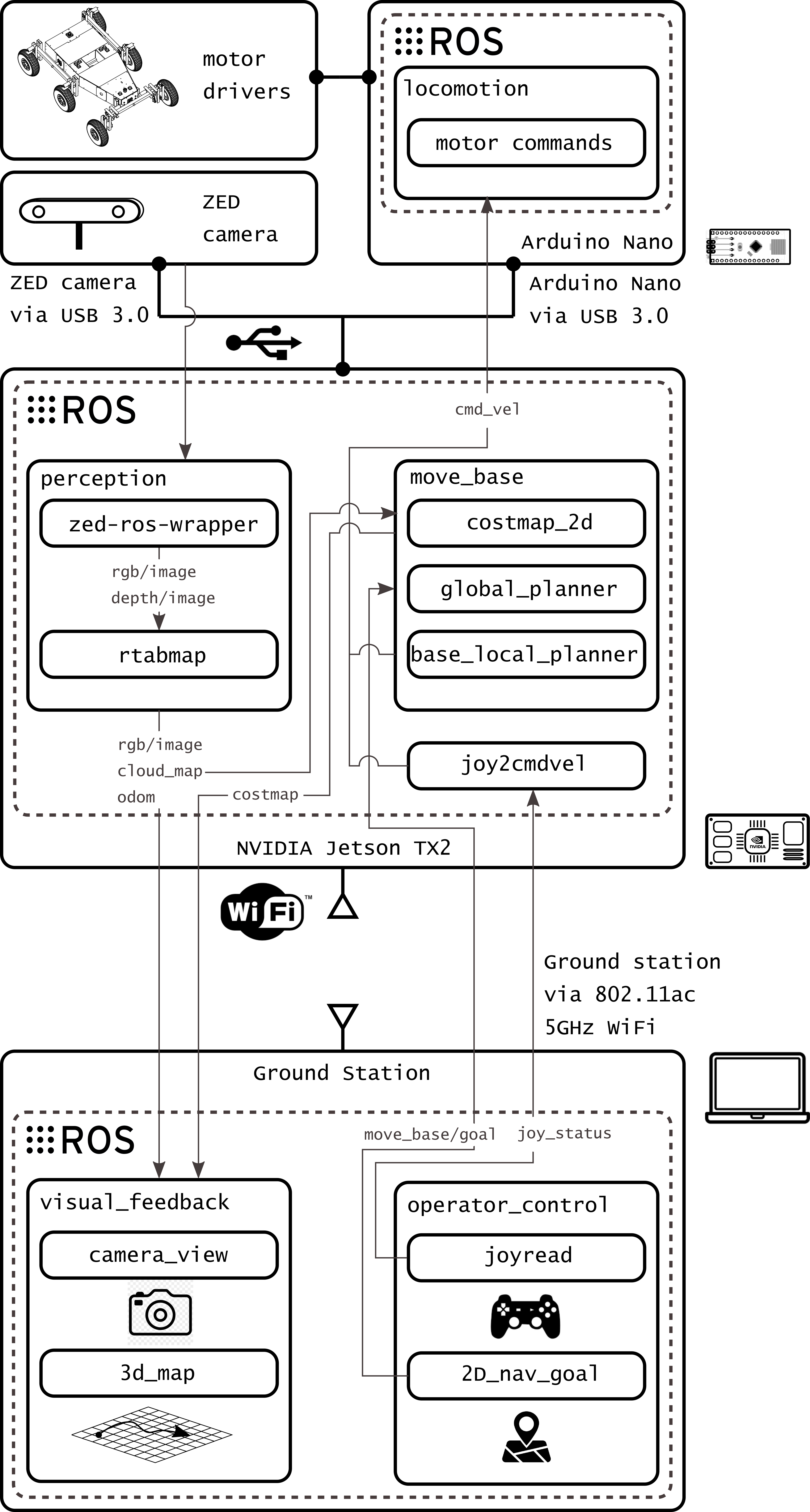}

\caption{System overview.}
\label{fig::system}
\end{figure}

\subsection{Perception}

\begin{figure}[!h]
\centering
\includegraphics[width=\linewidth]{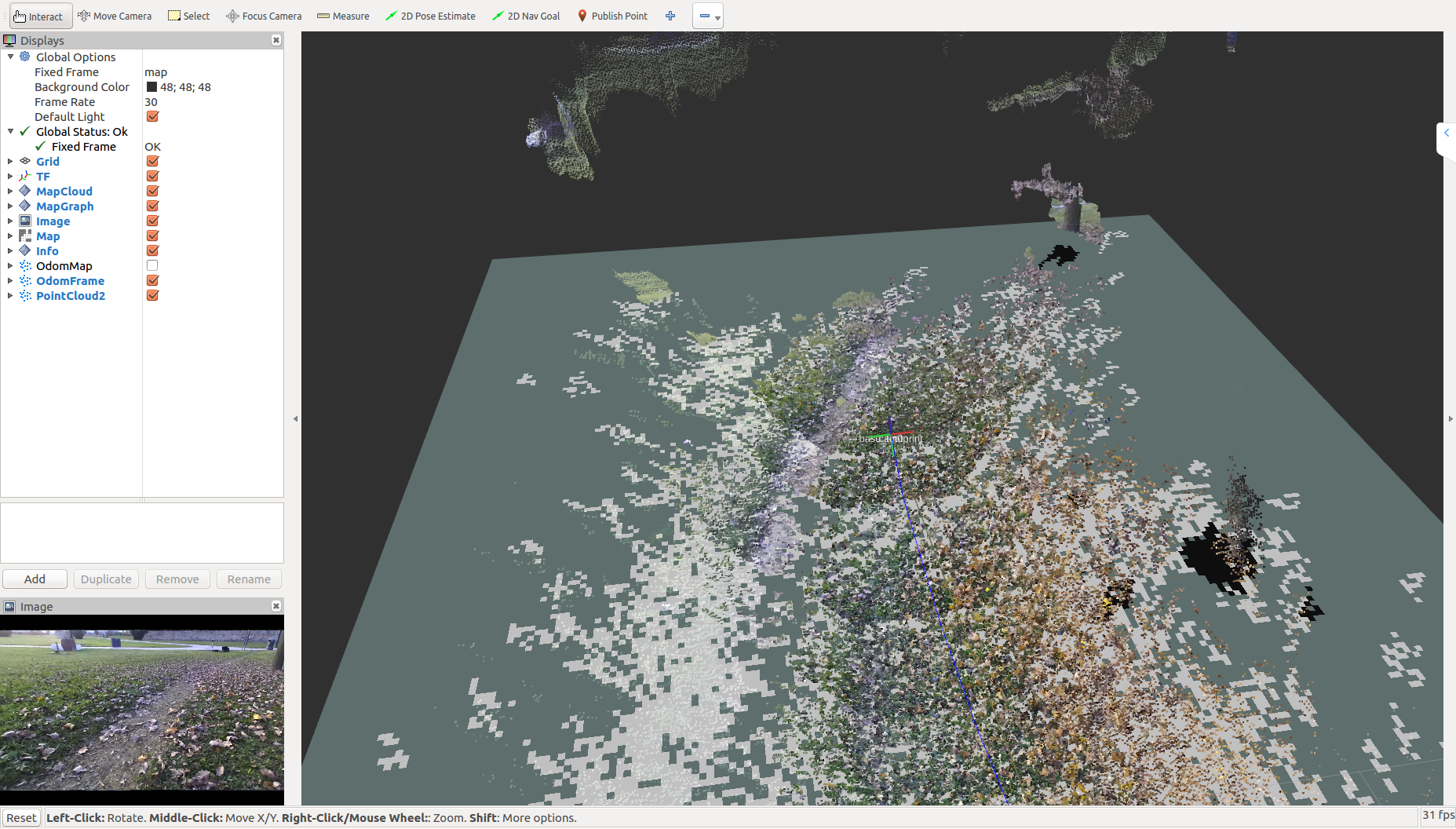}

\caption{Centre image: 3D point cloud of the environment obtained with RTAB-MAP algorithm, the 2D occupancy grid of potential obstacles and the current rover pose. Bottom left: current of the scene captured by rover's camera.}
\label{fig::explorationMap}
\end{figure}

Rover perception is based on RTAB-MAP \cite{labbe2013}, a standard feature-based front end visual SLAM algorithm handling stereo and RGB-D cameras. Loop closures are found using again a bag-of-words approach but it is offered a significant choice of features and descriptors. Multi-session mapping is provided by this algorithm, since it gives the possibility to save a map and a trajectory and to load them in a second time to resume the reconstruction. In \cite{giubilato2019} we have evaluated the performances of the most famous solutions for Visual SLAM compatible with the ROS middleware. RTAB-MAP has been chosen for its performances and because it provided a dense 3D map of the environment. To have a more accurate representation of the environment it is possible to enhance the sensing elements, as an example by adding depth sensor such as LiDARs or Time-of-Flight (ToF) cameras \cite{chiodini2020}.
The \verb!rtabmap! node take as input the ZED stereo images, connected to ROS via the \verb!zed-ros-wrapper!, and gives as output the odometry of the rover (\verb!odom!) and the 3D point cloud of the sensed environment (\verb!cloud_map!). \figurename{\ref{fig::explorationMap}} shows the point cloud map generated by means of the visual SLAM code RTAB-MAP.

\subsection{Navigation}

The navigation stack is based on the ROS navigation stack\footnote{http://wiki.ros.org/navigation}. It takes as input the rover’s odometry information, the range sensor streams, and the goal pose to outputs the velocity commands that are sent to the locomotion subsystem. It is composed by:
\begin{itemize}
    \item \verb!move_base!. This package links together a global and local planner to accomplish the global navigation task.
    \item \verb!costmap_2d!. This node uses as input the sensor data form the world, builds a 2D occupancy grid, and inflates costs in a 2D costmap based on the occupancy grid and a user specified inflation radius. In the proposed system the depth data retrieved from the stereo-camera are interpreted by means of an inverse measurement model into cells of the cost map which are occupied or free. In the used inverse measurement model, see \cite{andert2009drawing}, there is a distance measurement for each ray corresponding to the pixel of the depth map. The ray starts from the center of the camera and intersects the obstacle, starting from the camera the cells are free, then occupied at the obstacle and unknown after the obstacle.

    \figurename{\ref{fig::occupancy}} shows the occupancy map generated by the rover stereo camera. The 2D occupancy grid is built by cutting the 3D point cloud generated by the perception node at a height equal to 20 cm, which represents an insurmountable height for the rover. The cost map resolution has been set to 0.1 m, and its update frequency has been set to 4 Hz.
    \item \verb!global_planner!. This node implements the Dijkstra’s algorithm \cite{dijkstra1959} to find the minimum cost path from the rover’s starting pose to the rover’s goal pose. 
    \item \verb!base_local_planner!. This node implements a local trajectory planner based on the "Timed-Elastic-Band" approach \cite{rossmann2013}. This local planner receives as input the initial trajectory generated by the global planner and optimizes it during runtime with reference to minimize the trajectory execution time (time-optimal objective), separation from obstacles and compliance with kinodynamic constraints such as satisfying maximum velocities and accelerations. A differential drive kinematic model has been assumed. The maximal forward velocity has been set to 0.1 m/s, and the maximal acceleration has been set to 0.3 m/s$^2$.
\end{itemize}

\subsection{Locomotion}

Motor commands are converted to an USB serial protocol and sent to Arduino microcontrollers which create PWM signals for motor controllers. 

\begin{figure}[!h]
\centering
\includegraphics[width=\linewidth]{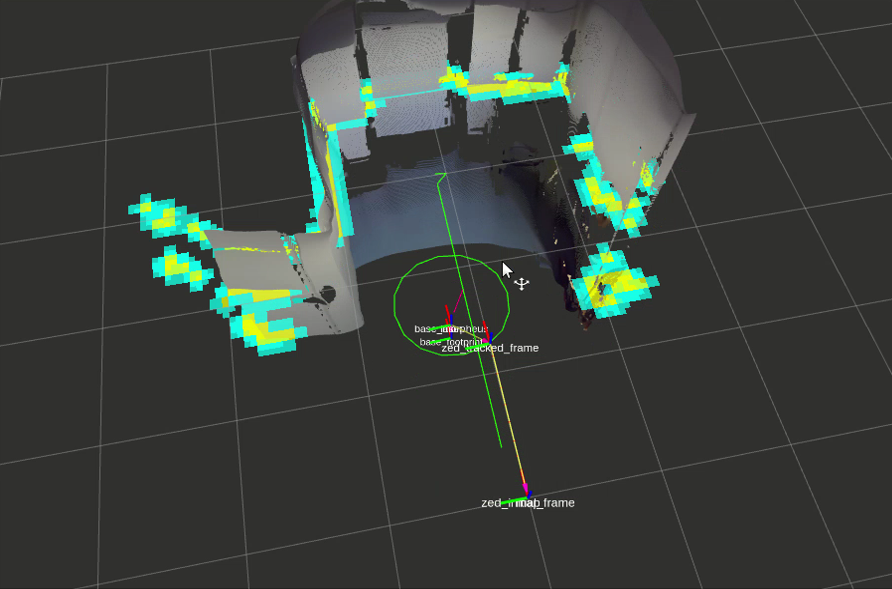}

\caption{3D view of the occupancy map generated and transmitted by the rover to the control base station. The yellow cells represent obstacles in the costmap, the cyan cells represent inflated obstacles, and the green polygon represents the footprint of the robot. For the robot to avoid collision, the footprint of the robot should never intersect a yellow cell and the center point of the robot should never cross a cyan cell.}
\label{fig::occupancy}
\end{figure}

\subsection{Communication}

The platform is remotely controlled by a ground control station using a WiFi connection. The on-board computer, a Jetson TX2 module has been configured to act as WiFi hot-spot. The base computer is connected to this WiFi hot-spot through the WiFi extender to establish a TCP connection.

\subsection{Operator Control}

There are two control modes: one involves sending commands via a PlayStation 3 joystick controller, the second method involves placing the rover's target pose directly on the streamed occupancy map.

\begin{itemize}
    \item \verb!joy2cmdvel!. This node transforms the button presses into velocity commands. The right lever is used to move forward and backward, and the left lever is used to rotate counterclockwise and clockwise. Moving and rotating speeds are proportional to the joystick lever position. The joystick inputs are handled by the node using the skid steering kinematics, the right and the left wheel banks speeds $V_l$ and $V_r$ are calculated as follow:
    \begin{equation}
    \begin{split}
        V_l & = V - \omega \ b/2\\
        V_r & = V + \omega \ b/2
    \end{split}
    \end{equation}
    where $V$ is the forward velocity, $\omega$ is the rotation velocity and $b$ is the distance between the rover wheels.
    \item \verb!2d_nav_goal!. This utility is used to give a goal position on the 3D view of the environment map displayed on the RViz viewer. 
\end{itemize}

\subsection{Visual Feedback}
The visualization tool is based on RViz (see \figurename{\ref{fig::visual_feedbak}}), which is the 3D visualization tool of ROS. The following messages are streamed via Wi-Fi and visualized:
\begin{itemize}
    \item \verb!camera_view!. Current view of the navigation camera.
    \item \verb!3d_map!. 3D map containing: the 3D point cloud of the environment obtained with the visual SLAM algorithm, the 2D occupancy grid of potential obstacles, the online update of the planned rover’s path to reach the navigation goal, the travelled trajectory and the rover footprint. It is possible to interact with the map via mouse control, rotate it, translate it and zoom it.
\end{itemize}

\begin{figure}[!h]
\centering
\includegraphics[width=\linewidth]{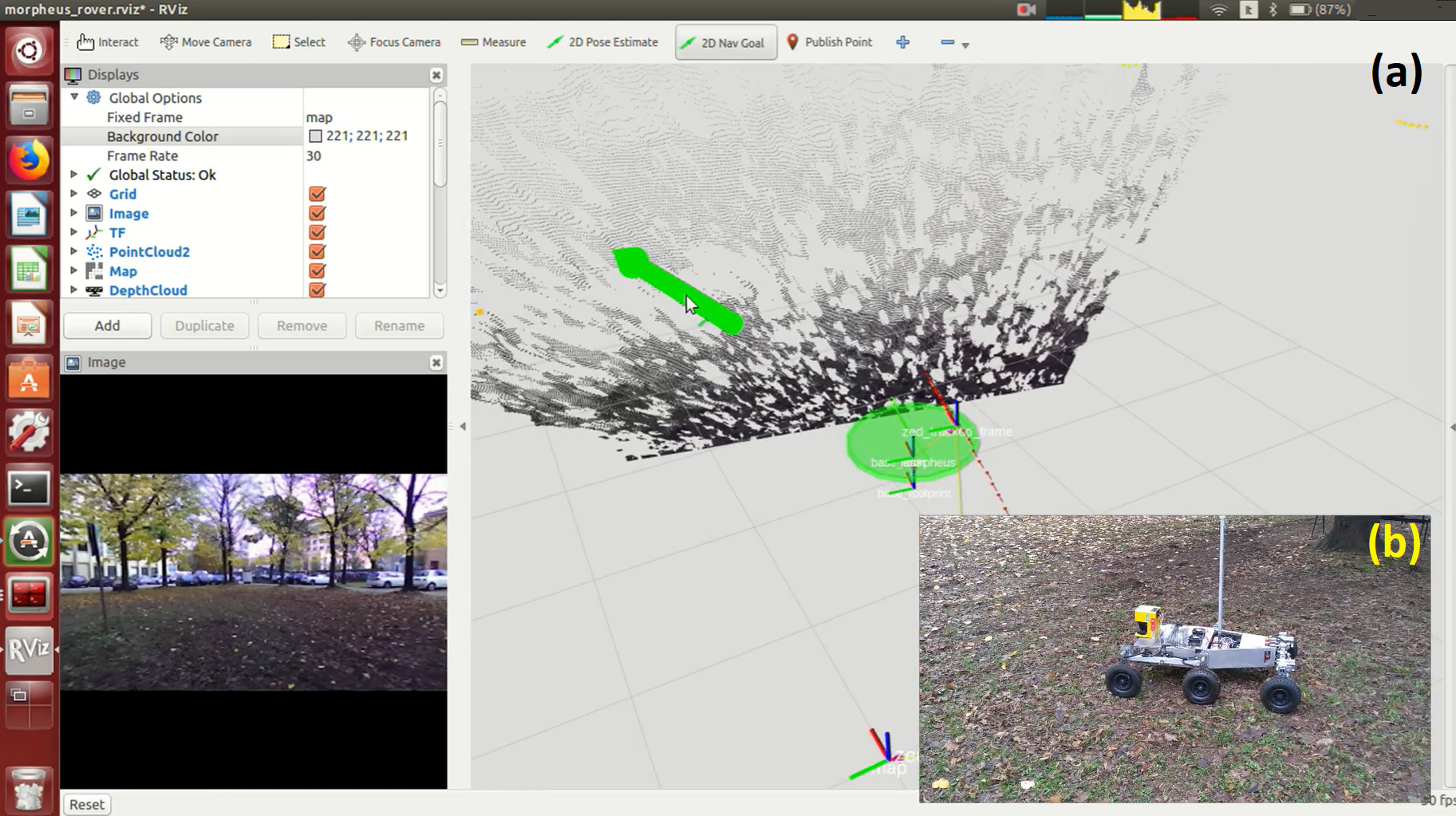}

\caption{(a) Rover pose goal positioned on the navigation map transmitted by the rover and displayed via the RViz viewer. In the center it is shown the current rover pose and the map build thanks to the visual SLAM node. Green arrow represents the goal pose ($x$, $y$ and $\theta$) to be reached by the rover. Bottom left image shows the current camera view of the rover. (b) Rover platform during the exploration task.}
\label{fig::visual_feedbak}
\end{figure}

\section{Test and Results}
\label{sec::test}

Preliminary tests have been conducted in both outdoor and indoor environments. \figurename{\ref{fig::occupancy}} shows the occupancy map and the related point cloud captured during a test performed in a office environment and \figurename{\ref{fig::visual_feedbak}} shows the rover operating in open field. At the following link it is possible to see a video about an operator controlling the rover via joystick inputs: https://youtu.be/AcjXQtZfacE. The experiment consisted in making the rover perform a loop trajectory in a very restricted area. In the video it is possible to see the rover perform a straight line, a left turn, an obstacle avoidance, a straight line and a point turn to return to the initial position. The difficulty of the maneuver should be noted because the size of the operating area was slightly larger than the size of the rover.

At the following link it is possible to see the navigation goal settle with interactive monitoring, and the self-awareness capabilities of the system: https://youtu.be/NRzOkjgjyeU. The experiment consisted of having the rover perform a 180° heading change on its own. The operator simply defines in the iterative viewer of the base station, based on RViz, the target position of the rover with an arrow, the tip of the arrow represents the direction the rover should have at the end of the maneuver.The rover executes the maneuver based on the cost map built on-line with the stereo-camera and the path planner based on time-elastic band theory.

In order for the operator to have scene situational awareness the following data are streamed to the base station: rover trajectory, current stereo camera point cloud, map point cloud, 2D cost map, and image left compressed at JPEG quality (80\%) at 4 fps.
Table \ref{tab::DataSize} shows average data size streamed to the base station, the operator can experience a situational awareness of the scene with a total data amount of 39 Mb/s (for a 100 m travel), that could be compatible with the available bandwidth for the future lunar missions. As an example NASA’s Lunar Laser Communications Demonstration proved that optical signals can reliably send information from lunar orbit back to Earth at a speed of 633 Mb/s \cite{koziol2019phoning}.

\begin{table}[htbp]
\caption{Average data size streamed to the base station}
\begin{center}
  \begin{threeparttable}
\begin{tabular}{|c|c|}
\hline
 
\textbf{Topic} & \textbf{Data Size (Mb/s)} \\
\hline
\hline
Trajectory & 0.01\\
Stereo Camera Point Cloud & 29.49\\
Map Point Cloud & 6.86\\
2D Cost Map & 0.10\\
Image left (JEPG 80\% @ 4 fps) & 2.07\\
\hline
TOTAL & 38.53\\
\hline
\end{tabular}

   \begin{tablenotes}
      \small 
      \item *For a 100 m travel.
      \end{tablenotes}
\end{threeparttable}
\end{center}

\label{tab::DataSize}
\end{table}

\section{Conclusions}
\label{sec::conclusiond}

The realization of crewed modules orbiting a planet or on a planet will enable the possibility to control exploration rovers by astronauts in low latency mode, paving the way to enhanced exploration capabilities. In this paper, we present a low latency and user-friendly system, based on ROS, for rover control. The proposed system has the characteristics of a cyber-physical system as it is able to self-correct its path as it faces obstacles not previously identified by the operator. Preliminary tests have shown that even inexperienced operators are able to control the platform. Moreover, the amount of data transmitted for operator situational awareness does not exceed the bandwidth limits that will be available for future Lunar missions. 

\section*{Acknowledgement}
This work has been supported by Progetti Innovativi degli Studenti, University of Padova. We would also like to thank the Morpheus Team for the discussions and participation in the experiments.

\bibliographystyle{IEEEtran}
\bibliography{IEEEabrv,bare_conf}

\end{document}